\documentclass{article}

\PassOptionsToPackage{numbers,compress}{natbib}
\usepackage[preprint]{neurips_2026}

\usepackage[utf8]{inputenc}
\usepackage[T1]{fontenc}
\usepackage{hyperref}
\usepackage{url}
\usepackage{graphicx}
\usepackage{subcaption}
\usepackage{booktabs}
\usepackage[table]{xcolor}
\usepackage{array}
\usepackage{float}
\usepackage{placeins}
\usepackage{amsmath}
\usepackage{amssymb}
\usepackage{mathtools}
\usepackage{microtype}
\usepackage{nicefrac}
\title{Thinking in Text and Images: Interleaved Vision--Language Reasoning Traces for Long-Horizon Robot Manipulation}

\author{%
Jinkun Liu\\
Tsinghua University\\
\And
Haohan Chi\\
Tsinghua University\\
\And
Lingfeng Zhang\\
Tsinghua University\\
\And
Yifan Xie\\
Tsinghua University\\
\AND
YuAn Wang\\
Beijing Institute of Technology\\
\And
Long Chen\\
Xiaomi Group\\
\And
Hangjun Ye\\
Xiaomi Group\\
\And
Xiaoshuai Hao\\
Xiaomi Group\\
\And
Wenbo Ding\\
Tsinghua University\\
}

\newcommand{\method}{IVLR}
\newcommand{\trace}{IVLR-Trace}
\definecolor{ivlrhighlight}{gray}{0.92}
\newcolumntype{L}[1]{>{\raggedright\arraybackslash}p{#1}}
\newcolumntype{C}[1]{>{\centering\arraybackslash}m{#1}}

\begin{document}

\maketitle

\begin{abstract}
Long-horizon robotic manipulation requires plans that are both logically coherent and geometrically grounded. Existing Vision-Language-Action policies usually hide planning in latent states or expose only one modality: text-only chain-of-thought encodes causal order but misses spatial constraints, while visual prediction provides geometric cues but often remains local and semantically underconstrained. We introduce Interleaved Vision--Language Reasoning (IVLR), a policy framework built around \trace{}, an explicit intermediate representation that alternates textual subgoals with visual keyframes over the full task horizon. At test time, a single native multimodal transformer self-generates this global semantic-geometric trace from the initial observation and instruction, caches it, and conditions a closed-loop action decoder on the trace, original instruction, and current observation. Because standard robot datasets lack such traces, we construct pseudo-supervision by temporally segmenting demonstrations and captioning each stage with a vision-language model. Across simulated benchmarks for long-horizon manipulation and visual distribution shift, \method{} reaches 95.5\% average success on LIBERO, including 92.4\% on LIBERO-Long, and 59.4\% overall success on SimplerEnv-WidowX. Ablations show that both modalities are necessary: without traces, LIBERO-Long success drops to 37.7\%; text-only and vision-only traces reach 62.0\% and 68.4\%, while the full interleaved trace reaches 92.4\%. Stress tests with execution perturbations and masked trace content show moderate degradation, suggesting that the trace can tolerate local corruption and moderate execution drift, but remains limited under stale or incorrect global plans. Our results position interleaved vision-language traces as a scalable representation for explicit multimodal reasoning in robot policies, while identifying static, fully observed environments and initial planning latency as current limitations.
\end{abstract}

\section{Introduction}

\begin{figure*}[t]
    \centering
    \includegraphics[width=\textwidth]{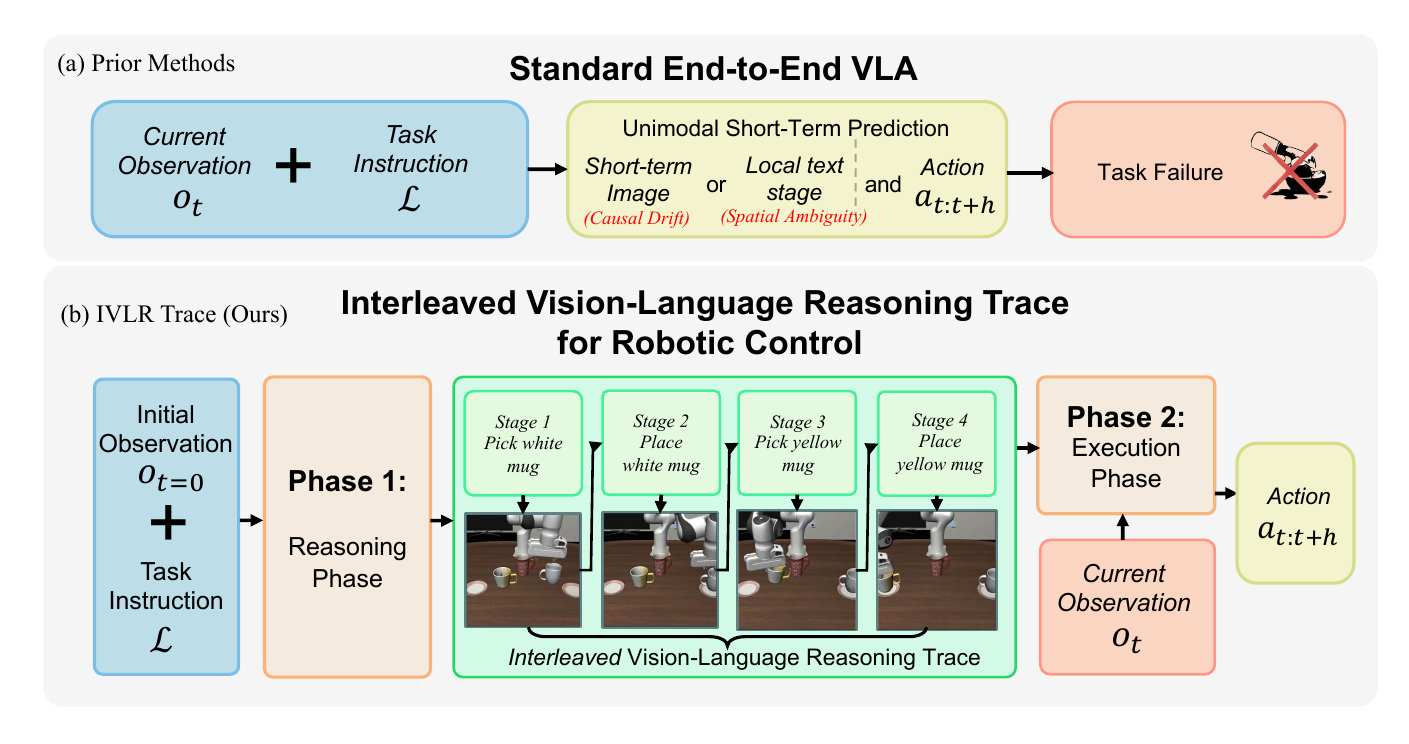}
    \vspace{-8pt}
    \caption{\textbf{Interleaved reasoning traces for long-horizon control.}
    Standard end-to-end VLA policies often predict actions or short local futures directly from the current observation and instruction. \method{} first generates a full-horizon \trace{} composed of textual subgoals and visual keyframes, then uses the cached trace together with the current observation and instruction for closed-loop action decoding.}
    \label{fig:teaser}
    \vspace{-8pt}
\end{figure*}

Long-horizon robotic manipulation is not only a problem of choosing the next motor command. A robot must preserve the causal order of a task while grounding each abstract subgoal in the geometry of the current scene. In a task such as placing a mug before manipulating a second object, the policy must remember which object matters first, where it should be placed, and how the scene should evolve after each intermediate interaction. These two requirements, causal coherence and geometric grounding, are difficult to satisfy with a purely reactive observation-language action mapping.

Recent Vision-Language-Action (VLA) policies have made rapid progress by scaling robot data and foundation-model backbones~\citep{brohan2022rt,zitkovich2023rt,kim2024openvla,black2024pi0,team2024octo,pertsch2025fast}. However, most policies still keep their long-horizon reasoning implicit in hidden states. This design is effective for many short-horizon tasks, but it gives the learner no explicit structure for representing intermediate task state, causal dependencies, or recoverable plans. As a result, when the instruction spans multiple stages, direct policies can become temporally myopic: they may greedily execute a visually salient later step or fail to maintain the state needed for a distant goal.

Explicit reasoning methods expose part of this hidden process, but usually through a single modality. Textual chain-of-thought and hierarchical language plans provide a useful scaffold for causal decomposition~\citep{ahn2022can,zawalski2024ecot,zhao2025cot,lin2025onetwovla}, yet language alone does not specify the spatial pose, contact geometry, or target placement needed by a low-level controller. Visual prediction and world-model policies provide geometric cues~\citep{hu2024vpp,cen2025worldvla,chen2025unified}, but local future frames can be semantically ambiguous: a plausible image need not be the correct step in the user's intended causal sequence. This complementarity motivates a representation that carries both modalities over the whole task.

We propose Interleaved Vision--Language Reasoning (IVLR), centered on \trace{}, an explicit semantic-geometric intermediate representation for long-horizon VLA control. A trace is a sequence of stages $z=[(c^{(1)}, k^{(1)}), \ldots, (c^{(N)}, k^{(N)})]$, where each textual subgoal $c^{(i)}$ records the causal role of a stage and each visual keyframe $k^{(i)}$ anchors the intended spatial state. Unlike a single caption, a text-only plan, or a one-step future prediction, \trace{} is a full-horizon storyboard. It makes the robot policy's intermediate reasoning visible, supervised, and directly ablatable.

We instantiate this representation in a single native multimodal transformer. At the beginning of an episode, the model generates the complete trace from the initial observation and instruction. During execution, the generated trace is cached as context; at every control step, the action decoder receives the current observation, original instruction, and cached trace. The policy is therefore not an open-loop replay of generated keyframes: it remains closed-loop with respect to the live observation, while using the trace as long-horizon semantic-geometric context. We do not impose an explicit stage pointer; instead, the current observation conditions which parts of the cached trace are relevant.

Standard robot demonstrations do not contain such interleaved reasoning traces. To obtain supervision, we retrofit demonstrations with pseudo-traces. We first segment each trajectory into stages using the Universal Visual Decomposer~\citep{zhang2024universal}, select the final frame of each segment as the visual keyframe, and then use a vision-language model to caption each stage. This pipeline is not intended to produce perfect symbolic plans. It provides scalable pseudo-supervision that lets us study whether full-horizon interleaved traces are useful for policy learning.

Our experiments focus on simulated manipulation benchmarks where long-horizon structure and visual distribution shift can be controlled. On LIBERO~\citep{liu2023libero}, \method{} achieves 95.5\% average success and 92.4\% on LIBERO-Long, outperforming the compared VLA and CoT baselines most clearly in the long-horizon suite. On SimplerEnv-WidowX~\citep{li2024evaluating}, it obtains 59.4\% overall success, compared with 42.7\% for SpatialVLA. Ablations show that the interleaving matters: on LIBERO-Long, removing traces gives 37.7\%, text-only traces give 62.0\%, visual-only traces give 68.4\%, and the full trace gives 92.4\%. Controlled stress tests further show that performance degrades moderately under 2 cm execution perturbations and 30\% trace masking, while remaining far above the no-trace baseline.

\paragraph{Contributions.}
First, we formulate long-horizon VLA control with an explicit \trace{}, a full-horizon sequence of textual subgoals and visual keyframes that represents semantic logic and geometric grounding in a single intermediate structure.
Second, we instantiate this representation in a native multimodal transformer that generates the trace once and then conditions closed-loop action decoding on the cached trace, original instruction, and current observation.
Third, we introduce a scalable pseudo-trace construction pipeline that retrofits unstructured robot demonstrations with staged vision-language supervision.
Finally, we show on simulated manipulation benchmarks that interleaved traces substantially improve long-horizon success, that both text and visual components are necessary, and that the trace-conditioned policy remains functional under controlled execution perturbations and trace corruptions.

\section{Related Work}

\paragraph{Vision-language-action policies.}
VLA models aim to bind visual perception, language instructions, and motor actions in a single policy. RT-1 and RT-2 showed that transformer-based robot policies can benefit from large-scale data and web-scale semantic knowledge~\citep{brohan2022rt,zitkovich2023rt}. OpenVLA, Octo, $\pi_0$, FAST, and related systems further improve openness, data mixture, action modeling, and execution speed~\citep{kim2024openvla,team2024octo,black2024pi0,pertsch2025fast,kim2025fine}. These policies typically predict actions directly, or compress task context into latent activations. \method{} instead exposes a full-horizon reasoning state before execution and makes it available as cached context for closed-loop control.

Table~\ref{tab:conceptual} previews the conceptual distinction. \method{} should not be viewed as merely adding visual tokens to CoT: its specific design choice is to generate a full-horizon semantic-geometric storyboard before execution and then reuse it as cached context for closed-loop control.

\begin{table}[H]
    \caption{Conceptual comparison with related reasoning and VLA families.}
    \label{tab:conceptual}
    \centering
    \footnotesize
    \setlength{\tabcolsep}{2.5pt}
    \renewcommand{\arraystretch}{1.08}
    \begin{tabular}{@{}C{0.18\textwidth}C{0.16\textwidth}C{0.24\textwidth}C{0.17\textwidth}C{0.17\textwidth}@{}}
        \toprule
        \rowcolor{black!6}
        \textbf{Method family} & \textbf{Horizon} & \textbf{Representation} & \textbf{Grounding} & \textbf{Execution} \\
        \midrule
        \rowcolor{black!2}
        Text CoT / SayCan-style & Global or hierarchical & Text plans or rationales & Indirect visual grounding & Planner--policy coupling \\
        Visual prediction / world models & Local or receding & Future frames or latent states & Strong visual, weak semantic & Recomputed or local \\
        \rowcolor{black!2}
        CoT-VLA & Local or receding & Visual tokens & Intermediate perceptual cues & Action-conditioned \\
        dVLA / UniVLA / EO-1 & Local or receding & Visual/Textual tokens & Varies by model & Unified generation \\
        \rowcolor{ivlrhighlight} \textbf{\method{} (ours)} & \textbf{Full horizon before execution} & \textbf{Text subgoals + RGB keyframes} & \textbf{Generated RGB anchors} & \textbf{Cached trace + closed-loop action} \\
        \bottomrule
    \end{tabular}
\end{table}

\paragraph{Textual reasoning for robots.}
Language-based robot planners decompose instructions into executable subgoals, code, affordance-scored actions, or chain-of-thought rationales~\citep{ahn2022can,huang2023grounded,stone2023open,zawalski2024ecot,zhao2025cot,lin2025onetwovla}. Their strength is causal abstraction: text can name objects, describe order, and explain intent. Their weakness is that text does not by itself resolve the detailed geometry of manipulation. \trace{} keeps this causal scaffold but pairs every stage with a visual keyframe, giving the action decoder a spatial anchor rather than a purely linguistic instruction.

\paragraph{Visual prediction and world models.}
Visual predictive policies and world-model VLAs learn to imagine future observations or latent dynamics~\citep{hu2024vpp,cen2025worldvla,chen2025unified}. These methods are naturally grounded in geometry and can represent contact-relevant scene changes. However, much of this prediction is local or receding-horizon, and visual futures without language can drift toward plausible but task-incorrect states. In long-horizon manipulation, this ambiguity matters because a visually feasible future may correspond to the wrong subgoal order. \method{} differs by generating a full-horizon interleaving of semantic captions and RGB keyframes, so the predicted visual states are constrained by the intended task sequence rather than only by short-term visual plausibility.

\paragraph{Unified multimodal generation.}
Native multimodal models such as Chameleon, Transfusion, and Show-o/Show-o2 unify understanding and generation across text and images in a single transformer~\citep{team2024chameleon,zhou2025transfusion,showo,xie2025show}. Robotics work has begun to adapt this idea to action generation, including UniVLA, dVLA, EO-1, and related unified policies~\citep{wang2025unified,wen2025dvla,qu2025eo1}. The unified formulation is useful for \method{} because the same model can express language tokens, visual keyframes, and action-conditioned context without handing off between separate planners and controllers. Our contribution is not the unified backbone itself. We use such a backbone to study a specific robot reasoning representation: a full-horizon interleaved trace that is generated before execution and then cached for closed-loop action decoding.

\paragraph{Relation to concurrent structured VLA reasoning.}
Several recent systems expose intermediate reasoning or perceptual tokens. CoT-VLA emphasizes visual chain-of-thought for action reasoning~\citep{zhao2025cot}; dVLA and UniVLA unify multimodal and action token streams~\citep{wen2025dvla,wang2025unified}; EO-1 studies interleaved vision-text-action pretraining~\citep{qu2025eo1}; MolmoAct uses depth-aware perception tokens and editable spatial plans~\citep{lee2025molmoact}. Table~\ref{tab:conceptual} abstracts these systems by design axis rather than ranking them by capability. \method{} is distinguished along four dimensions: it generates a full-horizon trace rather than only local intermediate cues; it interleaves text subgoals with RGB visual keyframes rather than relying only on text, depth, points, or action tokens; it uses a single native transformer with a flow-matching visual head; and it caches the resulting semantic-geometric trace as context for closed-loop execution. This positions \trace{} as an explicit representation for VLA reasoning rather than only a new action decoder or tokenization scheme.

\section{Method}

\begin{figure*}[t]
    \centering
    \includegraphics[width=\textwidth]{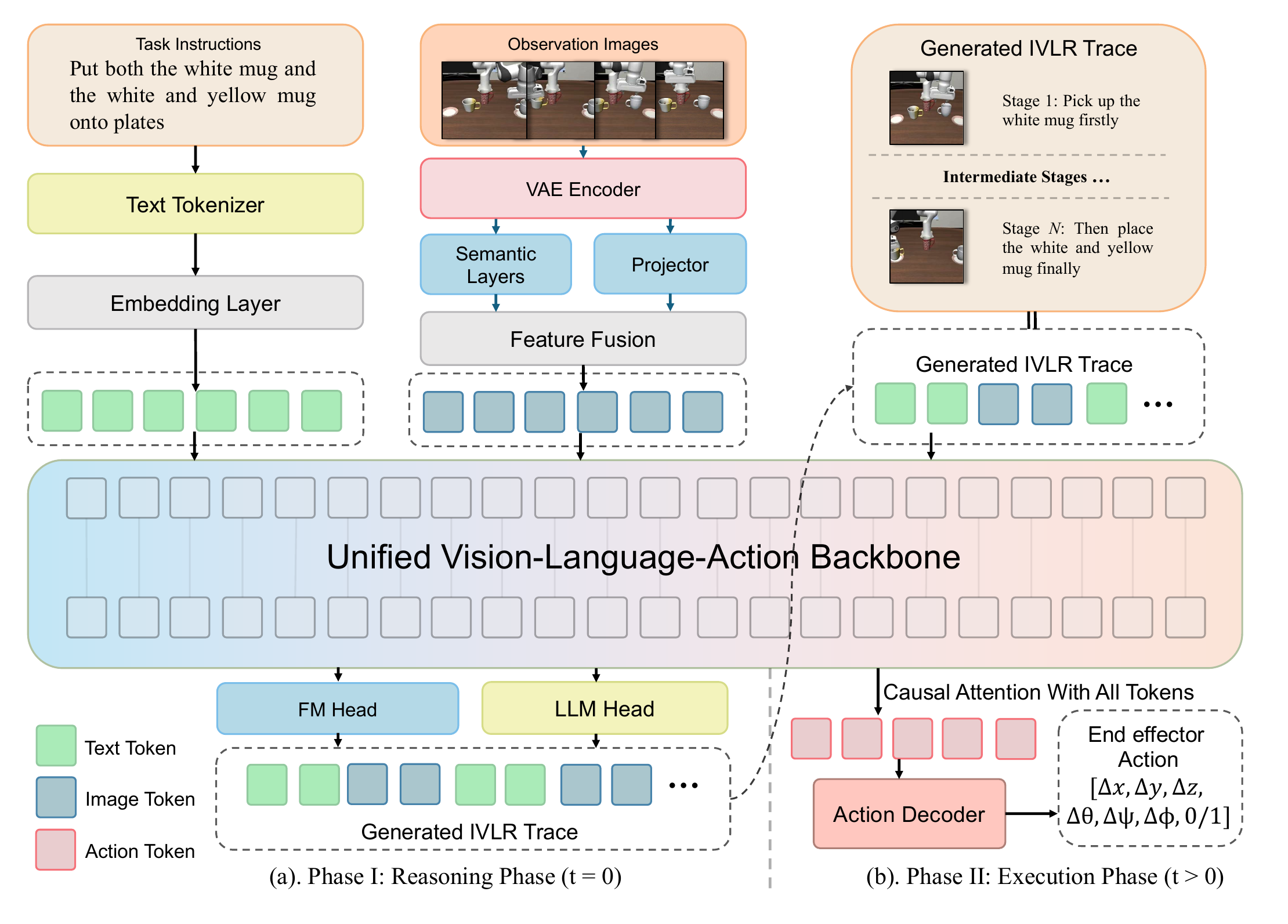}
    \vspace{-8pt}
    \caption{\textbf{Unified reasoning and execution architecture.}
    The same multimodal transformer first generates a full \trace{} from the initial observation and instruction. Textual subgoals are decoded by the language head and visual keyframes by a flow-matching visual head. During execution, the generated trace is cached and the action decoder predicts continuous actions from the current observation, instruction, and trace context.}
    \label{fig:architecture}
    \vspace{-8pt}
\end{figure*}

\subsection{Problem formulation}

We consider image-observation language-conditioned manipulation, where the full environment state is not directly observed. In the experiments studied here, however, the task-relevant workspace is assumed to be visible in the initial observation. At timestep $t$, the robot observes $o_t \in \mathcal{O}$, executes a continuous action $a_t \in \mathcal{A}$, and is conditioned on an instruction $\mathcal{I}$. A standard VLA policy learns a direct mapping $\pi_\theta(a_t \mid o_{t-h:t}, \mathcal{I})$, implicitly marginalizing over any planning structure needed for the task.

We introduce an explicit latent variable $z$, the \trace{}, generated from the initial observation and instruction:
\begin{equation}
    z = \left[(c^{(1)}, k^{(1)}), (c^{(2)}, k^{(2)}), \ldots, (c^{(N)}, k^{(N)})\right].
    \label{eq:trace_def}
\end{equation}
We call each pair $(c^{(i)}, k^{(i)})$ a semantic-geometric primitive: the caption specifies the causal role of the stage, while the keyframe specifies the expected visual state that makes the stage actionable. The policy factorizes reasoning and execution as
\begin{equation}
    P_\Theta(z, a_{1:T} \mid o_0, \mathcal{I}) =
    P_\Theta(z \mid o_0, \mathcal{I})
    \prod_{t=1}^{T} P_\Theta(a_t \mid o_t, \mathcal{I}, z).
    \label{eq:factorization}
\end{equation}
This factorization makes the intermediate plan explicit while keeping action prediction closed-loop in $o_t$. Although the trace already summarizes the instruction, we also keep the original instruction in the execution context to avoid losing task-level semantics when a generated trace is incomplete.

\subsection{Unified multimodal architecture}

We initialize from Show-o2 1.5B, a native multimodal transformer~\citep{xie2025show}, and adapt it to heterogeneous text, vision, and action representations.

\textbf{Text.}
Instructions and stage captions are represented with the language tokenizer and decoded autoregressively by the language head.

\textbf{Vision.}
Observations and keyframes are compressed with a Wan2.1 VAE~\citep{wan2025wan}. A visual frame $I$ is encoded as continuous latents $x=\mathrm{Flatten}(\mathcal{E}(I)) \in \mathbb{R}^{L \times d}$. For generated keyframes, the model uses a flow-matching head to predict the vector field $v_\Theta$ that transports noise to the target visual latent.

\textbf{Action.}
We introduce a learnable \texttt{[ACT]} token. During execution, its hidden state attends to the current observation, instruction, and cached trace, and a lightweight MLP predicts the continuous control:
\begin{equation}
    a_t = \mathrm{MLP}(h_{\texttt{[ACT]}}(o_t, \mathcal{I}, z)) + \mu_{\text{act}},
    \label{eq:action}
\end{equation}
where $\mu_{\text{act}}$ denotes dataset action statistics used for normalization.

\subsection{Reasoning and execution}

At $t=0$, the model generates a hypothesized global trace. It alternates between text decoding for $c^{(i)}$ and flow-matching visual generation for $k^{(i)}$. For visual generation, the model samples initial noise and solves the ordinary differential equation $dx_\tau / d\tau = v_\Theta(\tau, x_\tau, \mathcal{C})$, where $\mathcal{C}$ is the preceding multimodal context. The resulting latent is decoded to an RGB keyframe.

Once the trace is generated, it is cached. At every execution timestep, the model receives the live observation $o_t$, original instruction $\mathcal{I}$, and cached trace $z$, then predicts continuous actions using Eq.~\ref{eq:action}. We rely on learned implicit alignment rather than a hand-coded stage pointer: the action token attends to the full cached trace and live observation, allowing the model to select trace content compatible with the current state. We do not claim this alignment is perfectly interpretable; the perturbation and corruption studies in Section~\ref{sec:stress} serve as indirect tests of whether this mechanism remains functional under moderate drift.

\begin{figure*}[t]
    \centering
    \includegraphics[width=\textwidth]{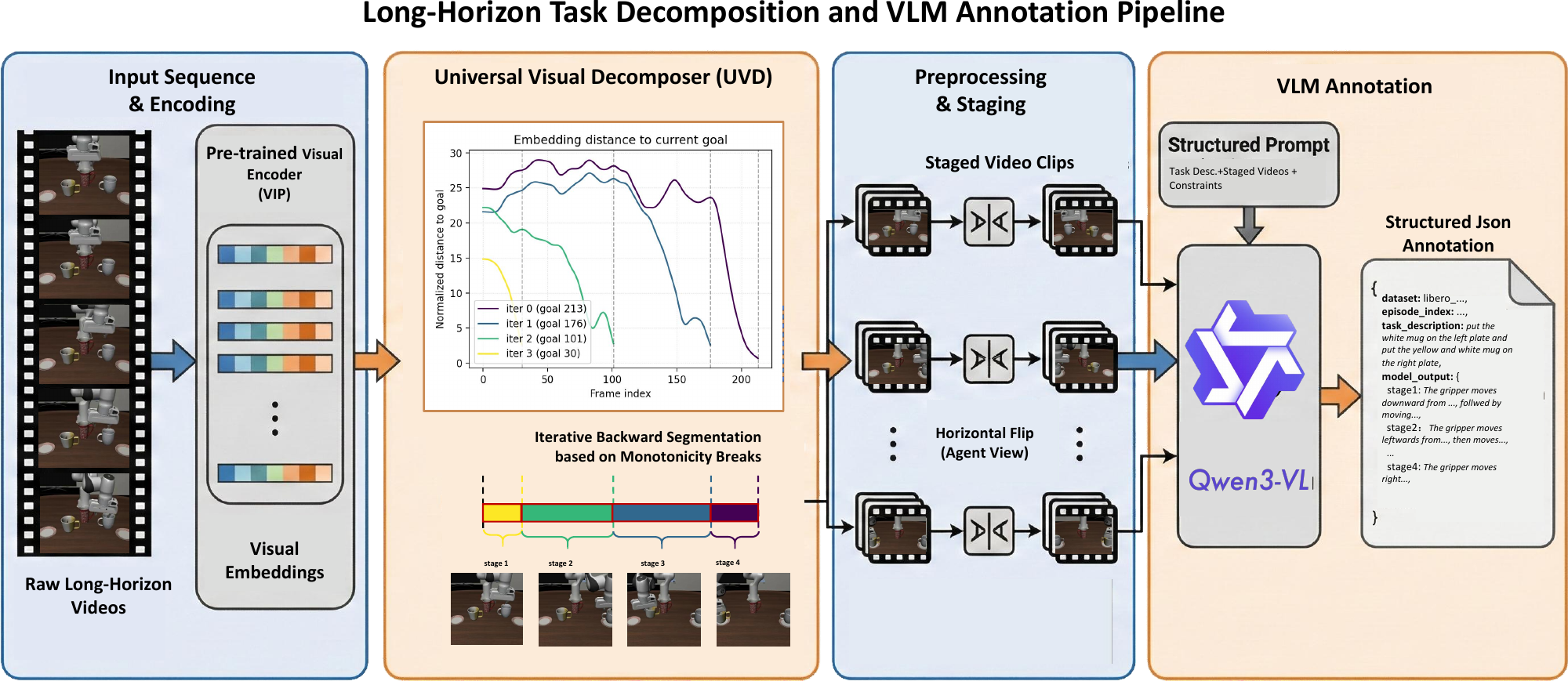}
    \vspace{-8pt}
    \caption{\textbf{Pseudo-trace construction.}
    Because standard robot datasets do not contain \trace{} annotations, we segment demonstrations into stages with UVD, choose segment endpoints as keyframes, and caption each stage with a VLM (Qwen3-VL in our implementation) to obtain pseudo-supervision.}
    \label{fig:pipeline}
    \vspace{-8pt}
\end{figure*}

\subsection{Pseudo-trace construction}

Training requires traces paired with demonstrations, but common robot datasets provide trajectories and instructions rather than staged multimodal reasoning. We therefore construct pseudo-traces automatically. Given a demonstration, we apply the Universal Visual Decomposer (UVD)~\citep{zhang2024universal} to identify stage boundaries from changes in visual embedding progress. The last frame of each segment becomes $k^{(i)}$. We then feed the segment and task context to a VLM, Qwen3-VL in our implementation~\citep{yang2025qwen3}, to produce a structured caption $c^{(i)}$ describing the relevant gripper-object interaction and spatial relation. Random manual inspection is used as a sanity check, but we treat these labels as pseudo-supervision rather than ground truth symbolic annotations; Appendix~\ref{app:pseudotrace} gives the annotation schema and filtering rules.

\subsection{Training objective}

Training mixes reasoning samples $(o_0,\mathcal{I},z_{\text{pseudo}})$ and action samples $(o_t,\mathcal{I},z_{\text{pseudo}},a_t)$. Reasoning samples supervise text and visual trace generation. Action samples supervise the continuous action decoder conditioned on trace context. To reduce over-reliance on perfect traces, we apply random trace noise and dynamic context masking during training. Since we do not ablate this training strategy separately, the stress tests in Section~\ref{sec:stress} evaluate the robustness of the resulting policy rather than isolating the causal effect of masking or noise injection.

The total objective is
\begin{equation}
    \mathcal{L}_{\text{total}} =
    \mathcal{L}_{\text{text}} +
    \lambda \mathcal{L}_{\text{flow}} +
    \gamma \mathcal{L}_{\text{action}}.
\end{equation}
$\mathcal{L}_{\text{text}}$ is next-token cross entropy on captions and text tokens. $\mathcal{L}_{\text{flow}}$ is the flow-matching loss for visual keyframes:
\begin{equation}
    \mathcal{L}_{\text{flow}} =
    \mathbb{E}_{\tau,x_0,x_1,\mathcal{C}}
    \left[\left\|v_\Theta(\tau,x_\tau,\mathcal{C})-(x_1-x_0)\right\|_2^2\right],
\end{equation}
where $x_1=\mathcal{E}(k^{(i)})$. $\mathcal{L}_{\text{action}}=\|\hat{a}_t-a_t\|_1$ is the L1 action loss.

\subsection{Efficiency setup}

The Show-o2-initialized model has 1.5B parameters. Generating the full trace takes approximately 10 seconds on a single NVIDIA H20 GPU. After the trace is cached, execution runs at 10 Hz with action chunking. Joint training uses 16 NVIDIA H200 GPUs; LIBERO training takes about 4 hours for 40K steps, and SimplerEnv training takes about 6 hours for 60K steps. This explicit reasoning design therefore trades an upfront thinking cost for higher long-horizon coherence during execution.

\FloatBarrier
\section{Experiments: When do interleaved traces help?}

We organize experiments around four questions. Does a full trace improve long-horizon control? Are text and visual components both necessary? Does the policy remain functional when execution or trace content is perturbed? What latency and compute trade-off does explicit reasoning introduce?

Unless otherwise stated, all \method{} results use self-generated traces at test time. Pseudo-traces constructed from demonstrations are used only as training supervision and are never provided as oracle plans during evaluation.

\subsection{Benchmarks}

\textbf{LIBERO.}
LIBERO contains four simulated manipulation suites that test spatial generalization, object variation, goal variation, and long-horizon sequential manipulation~\citep{liu2023libero}. We emphasize LIBERO-Long because it directly stresses causal ordering across multiple subgoals.

\textbf{SimplerEnv-WidowX.}
SimplerEnv evaluates robot policies in simulation environments designed to reflect visual distribution shifts such as lighting, background, and viewpoint variation~\citep{li2024evaluating}. We use it as a proxy for visual robustness under domain shift, not as a substitute for real-robot validation.

\begin{table}[H]
    \caption{Success rates on LIBERO. \method{} achieves the highest average among the compared methods and the strongest result on LIBERO-Long.}
    \label{tab:libero}
    \centering
    \small
    \setlength{\tabcolsep}{7.5pt}
    \renewcommand{\arraystretch}{1.08}
    \begin{tabular}{lccccc}
        \toprule
        Method & Spatial & Object & Goal & Long & Average \\
        \midrule
        DP*~\citep{chi2025diffusion} & 78.3\% & 92.5\% & 68.3\% & 50.5\% & 72.4\% \\
        Octo~\citep{team2024octo} & 78.9\% & 85.7\% & 84.6\% & 51.1\% & 75.1\% \\
        OpenVLA~\citep{kim2024openvla} & 84.9\% & 88.4\% & 79.2\% & 53.7\% & 76.5\% \\
        SpatialVLA~\citep{qu2025spatialvla} & 88.2\% & 89.9\% & 78.6\% & 55.5\% & 78.1\% \\
        CoT-VLA~\citep{zhao2025cot} & 87.5\% & 91.6\% & 87.6\% & 69.0\% & 81.1\% \\
        $\pi_0$-FAST~\citep{pertsch2025fast} & 96.4\% & \textbf{96.8\%} & 88.6\% & 60.2\% & 85.5\% \\
        VLA-0~\citep{goyal2025vla} & 97.0\% & 97.8\% & 96.2\% & 87.6\% & 94.7\% \\
        \midrule
        \rowcolor{ivlrhighlight} \textbf{\method{} (ours)} & \textbf{97.8\%} & 95.8\% & \textbf{97.0\%} & \textbf{92.4\%} & \textbf{95.5\%} \\
        \bottomrule
    \end{tabular}
\end{table}

\begin{table}[H]
    \caption{Success rates on SimplerEnv-WidowX. The overall success for \method{} is 59.4\%.}
    \label{tab:simpler}
    \centering
    \small
    \setlength{\tabcolsep}{4.2pt}
    \renewcommand{\arraystretch}{1.06}
    \resizebox{\textwidth}{!}{
    \begin{tabular}{lccccccccc}
        \toprule
        & \multicolumn{2}{c}{Spoon} & \multicolumn{2}{c}{Carrot} & \multicolumn{2}{c}{Stack} & \multicolumn{2}{c}{Eggplant} & \\
        \cmidrule(lr){2-3}\cmidrule(lr){4-5}\cmidrule(lr){6-7}\cmidrule(lr){8-9}
        Model
        & Grasp & Success & Grasp & Success & Grasp & Success & Grasp & Success & Success \\
        \midrule
        RT-1-X~\citep{zitkovich2023rt} & 16.7\% & 0.0\% & 20.8\% & 4.2\% & 8.3\% & 0.0\% & 0.0\% & 0.0\% & 1.1\% \\
        Octo-Base~\citep{team2024octo} & 34.7\% & 12.5\% & 52.8\% & 8.3\% & 31.9\% & 0.0\% & 66.7\% & 43.1\% & 16.0\% \\
        Octo-Small~\citep{team2024octo} & 77.8\% & 47.2\% & 27.8\% & 9.7\% & 40.3\% & 4.2\% & 87.5\% & 56.9\% & 29.5\% \\
        OpenVLA~\citep{kim2024openvla} & 4.1\% & 0.0\% & 33.3\% & 0.0\% & 12.5\% & 0.0\% & 8.3\% & 4.1\% & 1.0\% \\
        RoboVLMs~\citep{liu2025towards} & 70.8\% & 45.8\% & 33.3\% & 20.8\% & 54.2\% & 4.2\% & 91.7\% & 79.2\% & 37.5\% \\
        SpatialVLA~\citep{qu2025spatialvla} & 20.8\% & 16.7\% & 29.2\% & 25.0\% & 62.5\% & 29.2\% & \textbf{100\%} & \textbf{100\%} & 42.7\% \\
        \rowcolor{ivlrhighlight} \textbf{\method{} (ours)} & \textbf{75.0\%} & \textbf{70.8\%} & \textbf{54.2\%} & \textbf{45.8\%} & \textbf{79.2\%} & \textbf{33.3\%} & 95.8\% & 91.7\% & \textbf{59.4\%} \\
        \bottomrule
    \end{tabular}}
\end{table}

\FloatBarrier

\subsection{Main results}

The main results establish where \method{} helps; the ablations isolate why it helps; the stress tests probe whether the trace interface remains usable when its assumptions are moderately violated.

Table~\ref{tab:libero} shows the main LIBERO results. \method{} reaches 95.5\% average success. The largest difference appears on LIBERO-Long: \method{} achieves 92.4\%, compared with 69.0\% for CoT-VLA and 87.6\% for VLA-0. The result supports the central hypothesis that a full-horizon semantic-geometric trace is especially useful when tasks require maintaining causal order across multiple stages. We avoid claiming dominance on every suite: for example, $\pi_0$-FAST is slightly higher on Object.
The gain is concentrated on the Long suite rather than uniformly across all suites, which is consistent with the intended role of \trace{} as a long-horizon reasoning interface.

Table~\ref{tab:simpler} reports SimplerEnv-WidowX. \method{} obtains 59.4\% overall success, improving over SpatialVLA's 42.7\%. This result is evidence for robustness under simulated visual distribution shift, not evidence of completed real-robot deployment. We leave physical robot validation to future work.

Appendix~\ref{app:qualitative} provides qualitative traces and execution trajectories. The generated keyframes are not used as an open-loop replay target; they act as visual anchors that make intended intermediate states available to the action decoder.

\subsection{Trace ablations}

\begin{table}[H]
    \caption{Trace ablations on LIBERO. The no-trace variant removes both trace generation and trace conditioning; text-only, vision-only, and full \method{} isolate the contribution of each trace modality.}
    \label{tab:ablation}
    \centering
    \footnotesize
    \setlength{\tabcolsep}{3.6pt}
    \renewcommand{\arraystretch}{1.08}
    \begin{tabular}{lccccc}
        \toprule
        Variant & Spatial & Object & Goal & Long & Avg. \\
        \midrule
        w/o Trace & 81.2\% & 82.9\% & 72.4\% & 37.7\% & 68.6\% \\
        Text-Only Trace & 82.5\% & 88.1\% & 81.4\% & 62.0\% & 78.5\% \\
        Vision-Only Trace & 88.9\% & 91.2\% & 85.5\% & 68.4\% & 83.5\% \\
        \midrule
        \rowcolor{ivlrhighlight} \textbf{Full \method{}} & \textbf{97.8\%} & \textbf{95.8\%} & \textbf{97.0\%} & \textbf{92.4\%} & \textbf{95.5\%} \\
        \bottomrule
    \end{tabular}
\end{table}

Table~\ref{tab:ablation} isolates the trace interface. Removing traces reduces LIBERO-Long success to 37.7\%. Text-only traces improve the long-horizon result to 62.0\%, showing that language helps encode order but still lacks spatial anchors. Vision-only traces reach 68.4\%, indicating that visual anchors help, but geometry alone does not preserve the complete semantic sequence. The full interleaved trace reaches 92.4\%, indicating that the two modalities are complementary rather than interchangeable. Appendix~\ref{app:qualitative} illustrates the same pattern qualitatively.
The no-trace variant uses the same Show-o2 1.5B initialization, observation encoder, action decoder, action chunking, training data, and action loss as \method{}, but removes both trace generation and trace conditioning.

\subsection{Robustness under controlled stress tests}
\label{sec:stress}

\begin{table}[H]
    \caption{Controlled stress tests on LIBERO. Execution perturbation applies one random 2 cm end-effector displacement per episode after trace generation and before subsequent closed-loop recovery. Text masking masks 30\% of textual trace tokens; visual masking masks 30\% of keyframes. Stage order is preserved in both masking settings.}
    \label{tab:stress}
    \centering
    \small
    \setlength{\tabcolsep}{7pt}
    \renewcommand{\arraystretch}{1.08}
    \begin{tabular*}{0.86\textwidth}{@{\extracolsep{\fill}}lccccc}
        \toprule
        Setting & Spatial & Object & Goal & Long & Avg. \\
        \midrule
        Base & 97.8\% & 95.8\% & 97.0\% & 92.4\% & 95.5\% \\
        \midrule
        2 cm perturbation & 94.2\% & 92.5\% & 93.6\% & 88.2\% & 92.1\% \\
        Absolute drop & 3.6\% & 3.3\% & 3.4\% & 4.2\% & 3.4\% \\
        \midrule
        Text mask 30\% & 94.5\% & 92.0\% & 93.5\% & 88.6\% & 92.2\% \\
        Visual mask 30\% & 91.2\% & 90.5\% & 91.8\% & 86.4\% & 90.0\% \\
        \bottomrule
    \end{tabular*}
\end{table}

The cached trace raises two robustness questions: what happens when execution drifts away from the imagined trace, and what happens when the trace itself is incomplete? Table~\ref{tab:stress} evaluates the first case by applying one random end-effector displacement of approximately 2 cm after trace generation in each episode. Average success decreases from 95.5\% to 92.1\%, and LIBERO-Long decreases from 92.4\% to 88.2\%. This is a moderate degradation rather than a collapse, suggesting that closed-loop observation conditioning can recover from small execution slips.

The same table evaluates trace corruption by randomly masking 30\% of textual trace tokens or 30\% of visual keyframes at inference time while preserving the remaining stage order. Text masking reduces average success to 92.2\%; visual masking reduces it to 90.0\%. The larger drop under visual masking indicates that visual anchors are particularly important for precise manipulation. These results evaluate the robustness of the trained trace-conditioned policy; they do not isolate the causal contribution of the masking/noise training strategy, which would require a separate training ablation. These stress tests do not prove robustness to arbitrary plan errors, object removal, or dynamic environment changes. They show that the trace interface tolerates local missing content and moderate execution perturbations in the evaluated simulated settings.

\subsection{Latency, compute, and failure modes}

The explicit reasoning phase introduces a real trade-off. On a single H20 GPU, full trace generation takes about 10 seconds before execution begins. Once cached, action decoding runs at 10 Hz. This is suitable for studying long-horizon reasoning in static tabletop settings, but the upfront delay is a limitation for dynamic human-shared environments. A possible future optimization is asynchronous plan-while-acting, where the first stage is generated quickly and later trace stages are generated while the robot executes earlier ones; we do not evaluate that strategy here.

Observed failure modes differ from the dominant no-trace failures. Standard direct policies often fail by causal drift, such as selecting the wrong object or executing a later subgoal too early. \method{} reduces this failure pattern, but it can still fail under severe occlusion, object slip or other large physical stochasticity, scene changes after planning, and unobserved targets at $t=0$; Appendix~\ref{app:failures} summarizes these cases. These failures indicate that \method{} mainly reduces causal-drift failures, but does not solve perception failure, stale plans, or partial observability.

\section{Discussion and limitations}

The results support a representation-level conclusion: exposing a full-horizon interleaved trace makes long-horizon VLA reasoning more controllable than leaving all intermediate structure in latent activations. The trace is not merely a plan and not merely a memory. It is a semantic-geometric interface that lets the policy condition each closed-loop action on both the live observation and a structured description of intended future states.

The scope is deliberately limited. We evaluate in simulated manipulation benchmarks, including SimplerEnv as a proxy for visual distribution shift, but we do not claim real-robot deployment. The method assumes a static, fully observed workspace when the trace is generated. The current stress tests cover missing local trace content and moderate execution displacement, but not adversarially wrong stage order, object removal, or scene changes that invalidate the global trace. If the scene changes after planning, if targets are initially unobserved, or if another agent intervenes, the cached trace may become stale. Future work should extend \method{} with sliding-window replanning or self-triggered replanning, enabling the trace to be revised in dynamic scenes, along with uncertainty-aware trace revision and real-robot validation.

The method also pays an upfront latency cost. The current implementation spends about 10 seconds generating a complete trace before acting, although subsequent execution runs at 10 Hz. This trade-off is acceptable for some static long-horizon manipulation settings and problematic for fast-changing scenes. Reducing or hiding this cost through pipelined generation is an important direction, but it is not part of the present empirical claims.

\section*{Impact statement}

This work studies explicit multimodal reasoning representations for robot policies. Potential benefits include more inspectable robot decision processes and better long-horizon manipulation in structured environments. Potential risks arise if trace-conditioned policies are deployed without real-time safety monitors, because generated visual anchors can become stale when environments change. The present system is evaluated in simulation and should not be treated as sufficient for uncontrolled physical deployment without hardware validation, safety constraints, and interruption mechanisms.

\bibliographystyle{plainnat}
\bibliography{references}

\appendix
\raggedbottom

\section{Pseudo-trace annotation details}
\label{app:pseudotrace}

The pseudo-trace pipeline takes a demonstration video and task instruction as input and returns an ordered list of stages. UVD produces temporal boundaries over the demonstration; for each segment, we use the final frame as the visual keyframe because it is the first stable observation after the segment's interaction has completed. The VLM annotator receives the instruction, the ordered segment frames, and the selected keyframe index, then outputs a structured JSON record.

\paragraph{Annotation prompt template.}
The VLM is instructed to act as a robotics video annotator: given the task instruction and one segmented clip, describe the manipulation stage in one concise caption, identify the relevant objects, and state the gripper-object interaction or spatial relation that makes the stage complete. The prompt asks the model to avoid speculation about unseen objects and to output only valid JSON.

\paragraph{JSON schema.}
Each annotation uses the following fields:
\begin{verbatim}
{
  "stage_id": integer,
  "caption": string,
  "objects": [string],
  "gripper_action": string,
  "spatial_relation": string,
  "keyframe_index": integer
}
\end{verbatim}
Captions are allowed to be gripper-centric when the gripper action is necessary for defining the stage, but they must also mention the task-relevant object or spatial relation.

\paragraph{Filtering and inspection.}
We reject annotations that are not valid JSON, have empty captions, duplicate adjacent stages without a state change, refer to objects not visible in the segment or not relevant to the instruction, or describe only camera motion/background changes. Camera-view normalization, including horizontal flipping when required by the dataset view convention, is applied before VLM captioning and recorded as preprocessing metadata. Random manual inspection is used as a sanity check for segmentation boundaries, object names, and whether captions describe object state rather than only low-level motion. These labels remain pseudo-supervision rather than ground-truth symbolic annotations.

\paragraph{Accepted and rejected patterns.}
Accepted captions describe completed stage states, such as ``grasp the mug by its handle'' or ``place the object inside the target container.'' Rejected captions include empty or purely visual descriptions such as ``the camera view changes,'' duplicated adjacent captions with no new interaction, or object hallucinations inconsistent with the instruction and segment frames.

\section{Implementation and evaluation details}
\label{app:implementation}

The model is initialized from Show-o2 1.5B and trained with AdamW. We use $\beta=(0.9,0.95)$, $\epsilon=10^{-8}$, optimizer weight decay $10^{-8}$, gradient accumulation of 1, and gradient clipping/max gradient norm of 1.0. The maximum training budget is 100 epochs or 100K optimization steps, with 5K warmup steps, save interval 5K, evaluation interval 100, and logging frequency 10. The scheduler is cosine decay with minimum learning rate $10^{-6}$.

Learning rates are module-specific: base learning rate $2.5\times10^{-5}$, Show-o2 backbone/interface learning rate $1.0\times10^{-5}$, and action model learning rate $1.0\times10^{-4}$. Loss scaling uses 1.0 for VLA/action training and 0.1 for VLM/reasoning supervision. No modules are frozen in the reported setting.

For compute, joint training uses 16 NVIDIA H200 GPUs. The LIBERO runs reported in the main text use about 40K steps and take approximately 4 hours; the SimplerEnv runs use about 60K steps and take approximately 6 hours. At evaluation time, all \method{} entries use self-generated traces. Baseline numbers in the comparison tables follow the cited papers or benchmark reports unless otherwise noted. We report aggregate success rates over the standard evaluation episodes used by each benchmark; we do not report multi-seed confidence intervals due to compute constraints.

\section{Failure mode summary}
\label{app:failures}

Representative failures fall into four categories. Severe occlusion prevents the live observation from matching the visual anchors, so \method{} can fail for the same perceptual reason as direct VLA policies. Object slip and other large physical stochasticity can move the scene outside the recovery range assumed by the cached keyframes. Scene changes after planning make the trace stale, while direct VLA policies may still react locally but lack global task order. Finally, unobserved targets at $t=0$ cannot be represented by valid initial keyframes and require search or replanning beyond the current setup.

\section{Qualitative examples}
\label{app:qualitative}

\begin{figure}[H]
    \centering
    \includegraphics[width=\textwidth]{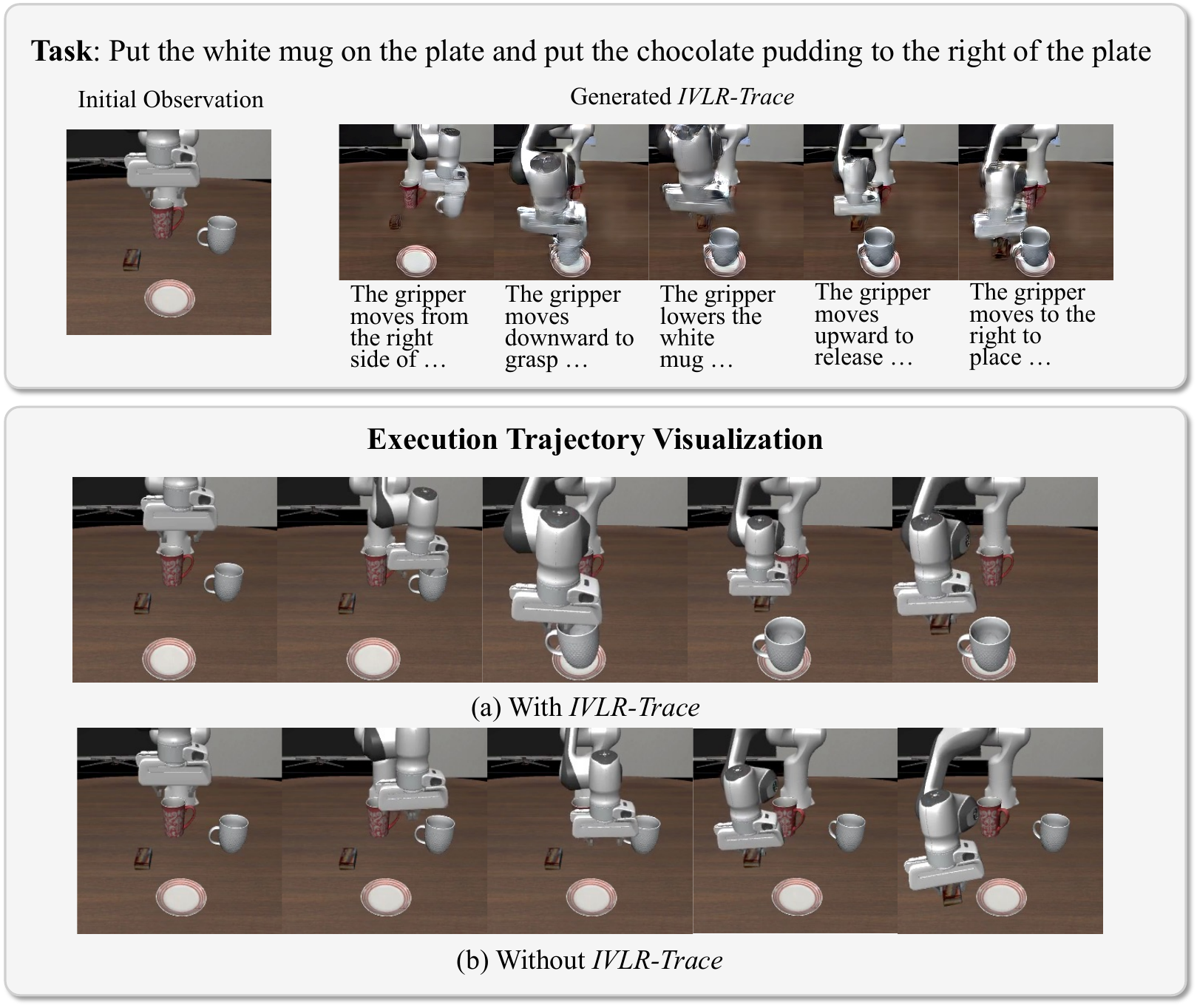}
    \caption{\textbf{Effect of explicit trace conditioning.}
    With a trace, the policy follows the intended causal order. Without the trace, the policy can greedily select a later visually salient object and fail the long-horizon task.}
    \label{fig:ablation}
\end{figure}

\begin{figure}[H]
    \centering
    \includegraphics[width=\textwidth]{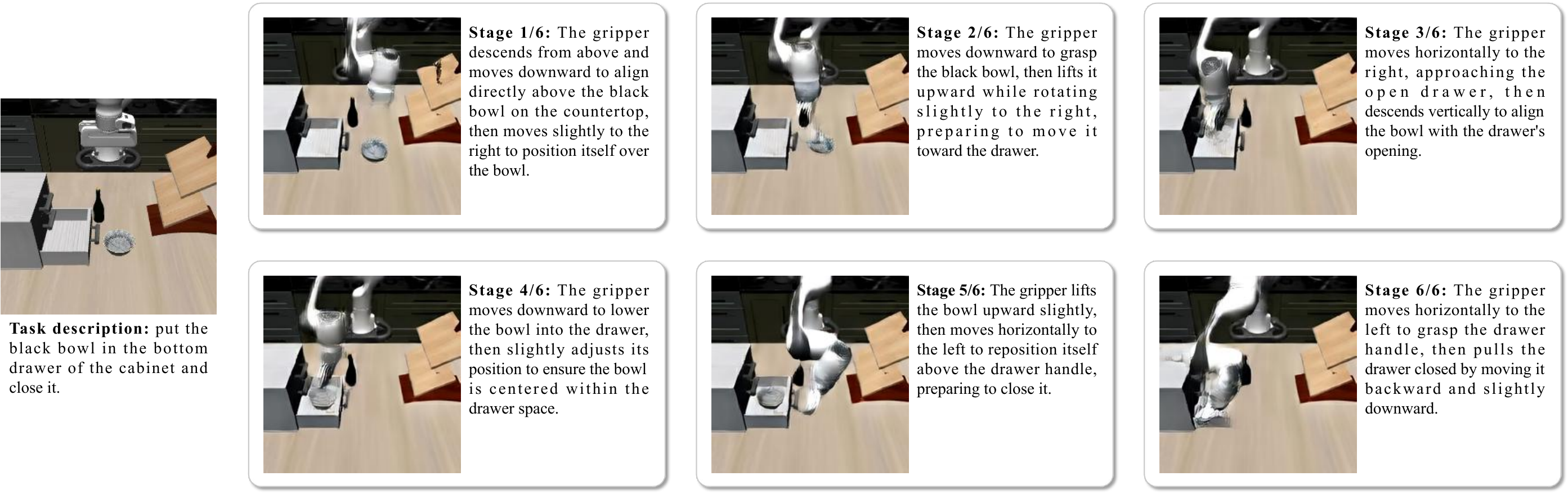}
    \caption{\textbf{Qualitative trace on LIBERO-Long.}
    The generated trace captures critical state transitions as interleaved captions and visual keyframes.}
    \label{fig:libero_vis}
\end{figure}

\end{document}